\documentclass[10pt,twocolumn,letterpaper]{article}
\usepackage[accsupp]{axessibility}

\usepackage{cvpr}              

\usepackage{graphics}
\usepackage[utf8]{inputenc} 
\usepackage[T1]{fontenc}    
\usepackage{url}            
\usepackage{booktabs}       
\usepackage{amsfonts}       
\usepackage{nicefrac}       
\usepackage{microtype}      
\usepackage[utf8]{inputenc}
\usepackage{graphicx}
\usepackage{amsmath}
\usepackage{amssymb}
\usepackage{mathtools}
\usepackage{amsthm}
\usepackage{arydshln}
\usepackage{multirow}
\usepackage{wrapfig, lipsum, booktabs}
\usepackage{algorithm}
\usepackage{enumitem}
\usepackage{paralist, tabularx}
\usepackage{balance}
\usepackage[noend]{algpseudocode}
\usepackage{pgfplots}
\usetikzlibrary{pgfplots.groupplots}
\pgfplotsset{compat=1.3}
\usepackage{tikz}
\usetikzlibrary{patterns}
\usepackage{pgf-pie}
\usepackage{adjustbox}
\usepackage{colortbl}

\newcommand{\impro}[1]{{\hspace{0.05cm}{\color[HTML]{32CB00}\textbf{(+#1)}}}}
\newcommand{\degra}[1]{{\hspace{0.05cm}{\textcolor{red}{\textbf{(-#1)}}}}}

\definecolor{demphcolor}{RGB}{144, 144, 144}
\definecolor{mygray}{gray}{0.4}
\definecolor{lightgray}{rgb}{0.9, 0.9, 0.9}
\newcommand{\demph}[1]{\textcolor{demphcolor}{#1}}

\definecolor{dt}{HTML}{ADCAD8}
\definecolor{dt2}{HTML}{cddfe7}

\newcommand{\modelname}{TimeChat\xspace}
\newcommand{\datasetname}{TimeIT\xspace}

\usepackage{pifont}
\definecolor{my_green}{RGB}{51,102,0}
\definecolor{my_red}{RGB}{204, 0, 0}
\newcommand{\cmark}{\textcolor{my_green}{\ding{51}}} 
\newcommand{\xmark}{\textcolor{my_red}{\ding{55}}} 

\DeclareMathAlphabet{\mathsfit}{\encodingdefault}{\sfdefault}{m}{sl}
\SetMathAlphabet{\mathsfit}{bold}{\encodingdefault}{\sfdefault}{bx}{n}

\definecolor{oorange}{RGB}{252,218,227}
\definecolor{yyellow}{RGB}{255,237,203}
\definecolor{ppurple}{RGB}{208,205,226}
\definecolor{ggreen}{RGB}{195,222,176}
\definecolor{ggrey}{RGB}{230,230,230}
\definecolor{rred}{RGB}{247,187,187}
\definecolor{wwhite}{RGB}{255,255,255}

\definecolor{cvprblue}{rgb}{0.21,0.49,0.74}
\usepackage[pagebackref,breaklinks,colorlinks,citecolor=cvprblue]{hyperref}

\def\paperID{14939} 
\def\confName{CVPR}
\def\confYear{2024}

\title{TimeChat: A Time-sensitive Multimodal Large Language Model\\ for Long Video Understanding}

\author{Shuhuai Ren$^{1}$\thanks{Equal contribution},\quad Linli Yao$^{1*}$,\quad Shicheng Li$^{1}$,\quad Xu Sun$^{1}$,\quad Lu Hou$^{2}$\\
$^{1}$National Key Laboratory for Multimedia Information Processing, 
\\School of Computer Science, Peking University\\
$^{2}$Huawei Noah's Ark Lab\\
\texttt{\{shuhuai\_ren, linliyao\}@stu.pku.edu.cn} \quad 
  \texttt{\{lisc99, xusun\}@pku.edu.cn} \\ \texttt{houlu3@huawei.com} \\
}

\begin{document}



\makeatletter
\let\@oldmaketitle\@maketitle%
\renewcommand{\@maketitle}{\@oldmaketitle%
\vspace{-2em}
    \centering
    \includegraphics[width=.9\linewidth]{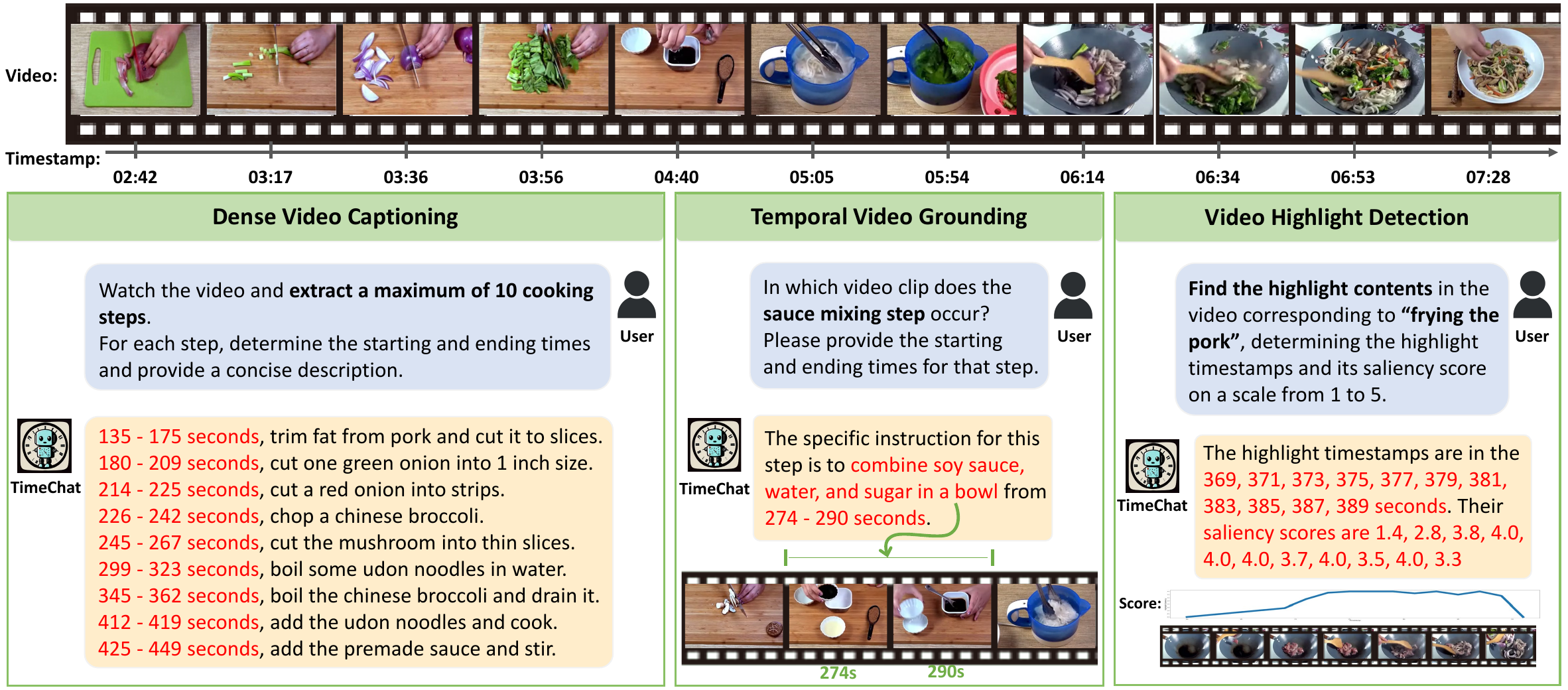}
    \captionof{figure}{\textbf{An illustration of temporal localization capability of \modelname.} 
    \modelname is a time-sensitive video large language model. Its capabilities extend beyond regular video captioning and Q\&A. Notably, \modelname can also follow user instructions to \textbf{(1)} summarize key events and pinpoint their moments in long videos (left block), \textbf{(2)} locate the start and end timestamps that correspond to user queries (middle block), and \textbf{(3)} detect highlight clips within the video (right block).
    }
    \label{fig:teaser}
   \bigskip}%
\makeatother

\maketitle

\begin{abstract}
This work proposes TimeChat, a time-sensitive multimodal large language model specifically designed for long video understanding. Our model incorporates two key architectural contributions: (1) a timestamp-aware frame encoder that binds visual content with the timestamp of each frame, and (2) 
a sliding video Q-Former that produces a video token sequence of varying lengths to accommodate videos of various durations. 
Additionally, we construct an instruction-tuning dataset, encompassing 6 tasks and a total of 125K instances, to further enhance TimeChat's instruction-following performance. Experiment results across various video understanding tasks, such as dense captioning, temporal grounding, and highlight detection, demonstrate TimeChat's strong zero-shot temporal localization and reasoning capabilities. For example, it achieves +9.2 F1 score and +2.8 CIDEr on YouCook2, +5.8 HIT@1 on QVHighlights, and +27.5 R@1 (IoU=0.5) on Charades-STA, compared to state-of-the-art video large language models, holding the potential to serve as a versatile video assistant for long-form video comprehension tasks and satisfy realistic user requirements.\footnote{Our code and dataset are available at \url{https://github.com/RenShuhuai-Andy/TimeChat}.}
\end{abstract}

\section{Introduction}
\label{sec:intro}
From educational tutorials to feature films, long-form videos have been an essential medium in our daily lives. 
However, it is both time-consuming and frustrating for individuals to sift through lengthy videos. Instead, human attention is consistently drawn to meaningful or highlighted visual segments such as essential steps in a cooking tutorial~\citep{Zhou2017TowardsAL} or fantastic moments from sports events~\citep{Lei2021QVHighlightsDM}. An intelligent time-sensitive video assistant to analyze long videos for users, encompassing temporal localization, timestamp detection, and key moment summarization, is a longstanding pursuit of the community.
With the emergence of Large Language Models (LLMs) and their impressive capacity to execute human instructions~\citep{chatgpt, vicuna2023, touvron2023llama, yao2023edit}, a natural question arises, i.e. Is it feasible to develop an LLM-based assistant for long-form video comprehension tasks to satisfy realistic user requirements?

Preliminary endeavors have been made to integrate video encoders with LLMs for basic video understanding including detailed captioning and question answering~\citep{Li2023VideoChatCV, Zhang2023VideoLLaMAAI, Luo2023ValleyVA, Song2023MovieChatFD, Liu2023VideoTellerEC, Maaz2023VideoChatGPTTD, jin2023chatunivi}. 
However, existing Video LLMs (VidLLMs) can only capture global visual semantics for short clips and fail to associate the significant video content with accurate timestamps.  
For example, Video-LLaMA~\citep{Zhang2023VideoLLaMAAI} and VideoChat~\cite{Li2023VideoChatCV} struggle to localize and describe meaningful events in untrimmed videos, leading to a low accuracy verified in Tab.~\ref{table:zeroshot}. 
Two main obstacles hinder the performance of existing VidLLMs. 
\textbf{Firstly}, their rigid compression converting video tokens to a fixed number (e.g. 32~\cite{Zhang2023VideoLLaMAAI}) is unsuitable for long-form video input~\cite {Ren2023TESTA}. It neglects the video's duration and results in severe spatial-temporal semantics degradation when processing massive frames from long videos. 
\textbf{Secondly}, they handle video and timestamp information separately without considering the explicit time-vision association thus being unable to localize timestamps accurately.


In this paper, we propose \textbf{\modelname}, a time-sensitive multimodal large language model for long video understanding and accurate temporal localization. 
To handle long video input, we propose a sliding video Q-Former to accommodate adaptive video token length during the extraction and compression of video features. Specifically, the video Q-Former compresses the frames within a sliding window into video tokens. By temporally moving the window, we can dynamically create a video token sequence of varying lengths to accommodate videos of various durations. It preserves the significant visual semantics of long videos and leads to more expressive and scalable video representation. 
Furthermore, to enhance the vision-timestamp association, we propose a time-aware frame encoder, which explicitly binds the visual context with the timestamp description of each frame. 


To stimulate the intrinsic timestamp localization capability of \modelname and enhance its instruction-following ability, 
we construct a novel instruction tuning dataset \textbf{\datasetname} involving diverse timestamp-related user instructions. 
This dataset is compiled from a variety of timestamp-associated long-video datasets with an average video length of 190.8 seconds. It is composed of 6 diverse tasks, 12 widely-used academic benchmarks, and a total of 125K instances. We reformat the original academic datasets into dialog style with manually written high-quality instructions. 
To our best knowledge, \datasetname is the first time-sensitive video-centric dataset designed for instruction tuning, with the aim to facilitate the development of VidLLMs. 

Utilizing the \datasetname dataset, We perform instruct tuning on \modelname and then assess its performance across various downstream tasks including dense video captioning, temporal video grounding, and video highlight detection. Experimental results show that our model substantially outperforms previous VidLLMs under zero-shot settings, with +9.2 F1 score and +2.8 CIDEr on YouCook2~\citep{Zhou2017TowardsAL}, +5.8 HIT@1 on QVHighlights~\citep{Lei2021QVHighlightsDM}, and +27.5 R@1 (IoU=0.5) on Charades-STA~\citep{Gao2017TALLTA}, respectively. Furthermore, qualitative results in new domains such as movie~\cite{Yue2023Movie101AN} and egocentric videos~\cite{Grauman2021Ego4DAT} demonstrate the generalization of \modelname towards a versatile and practical video assistant.


\begin{figure*}[t]
\centering
\includegraphics[width=0.88\textwidth]{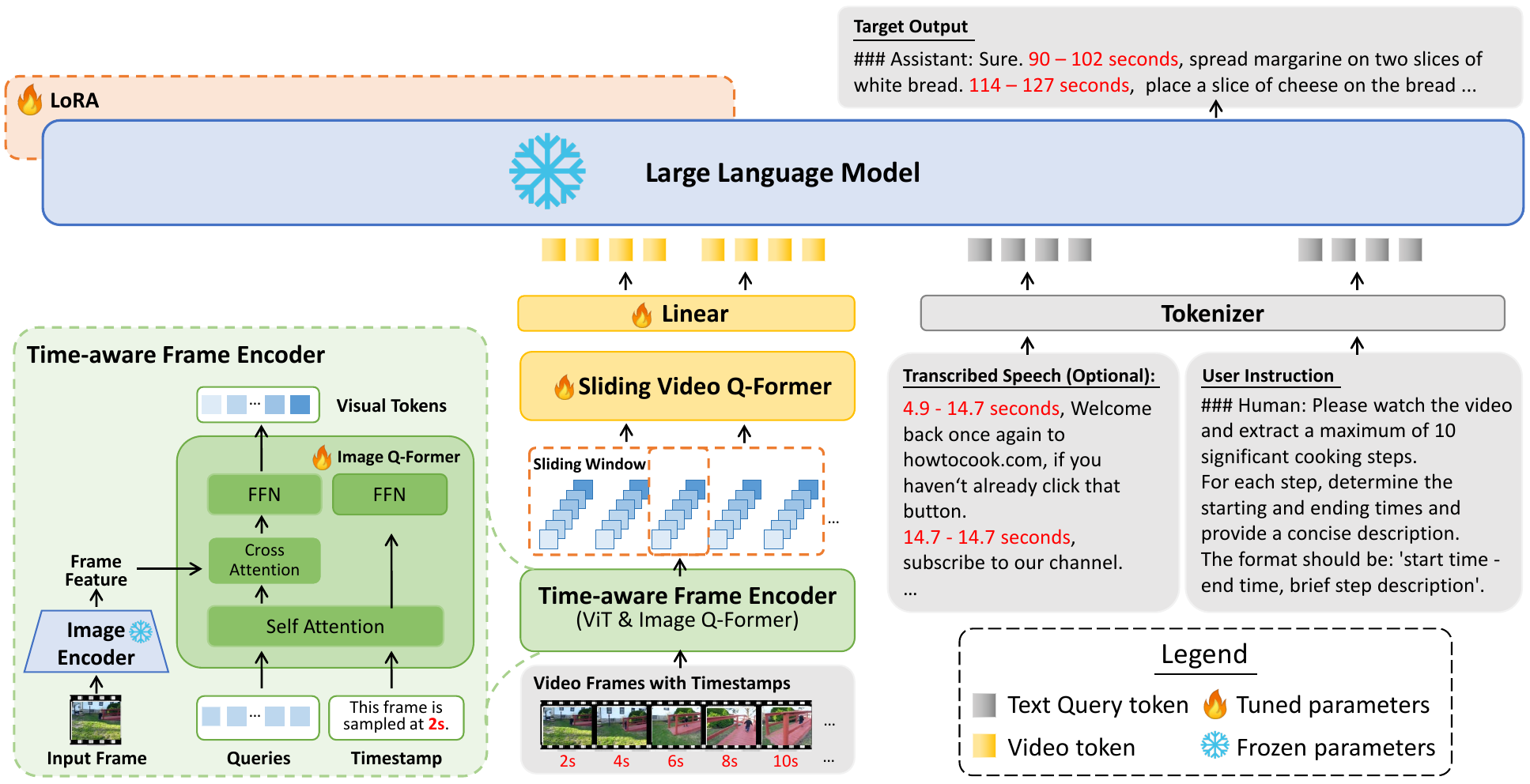}
\vspace{-2pt}
\caption{The overall architecture of \modelname.
Input a sequence of video frames along with their timestamps, \textbf{(a) Time-aware Frame Encoder} firstly extracts spatial tokens of each frame and binds the visual tokens with the corresponding timestamp description in frame-level. Then  \textbf{(b) Sliding Video Q-Former} establishes temporal relations across frame tokens with a moving sliding window which produces varied-length video tokens. Finally,  the video tokens are concatenated with the optional transcribed speech and the user query as input for a \textbf{(c) Large Language Model}, which produces appropriate responses.
}
\vspace{-5pt}
\label{fig:arch}
\end{figure*}

\section{Related Work}
\label{sec:related-work}

\subsection{Video Large Language Models}
With advancements in Large Language Models (LLMs), numerous studies have endeavored to integrate LLMs with a video encoder, thereby harnessing the powerful comprehension and generation capabilities of LLMs for video tasks~\citep{Li2023VideoChatCV, Zhang2023VideoLLaMAAI, Luo2023ValleyVA, Song2023MovieChatFD, Liu2023VideoTellerEC, Maaz2023VideoChatGPTTD, jin2023chatunivi}. 
These studies typically employ open-source LLMs such as Vicuna~\citep{vicuna2023} and LLaMA~\citep{touvron2023llama}. 
Their key difference lies in how they encode the video into vision tokens compatible with the LLMs.
Representative work like VideoChat~\citep{Li2023VideoChatCV} utilizes a video transformer to encode video features and subsequently implement a Query Transformer (Q-Former)~\citep{Li2023BLIP2BL} to compress video tokens. Video-LLaMA~\citep{Zhang2023VideoLLaMAAI} first uses a vision transformer (ViT)~\citep{Dosovitskiy2020AnII} with an image Q-Former to encode individual frames, and then employs a video Q-Former for temporal modeling. However, these methods compress video tokens to a fixed number, resulting in visual semantic degradation when handling lengthy videos.
In contrast, our model \modelname offers adjustable compression rates for visual tokens, increasing adaptability to varying video lengths. Moreover, our model explicitly establishes a frame-level vision-timestamp relationship to improve temporal localization capabilities.

\subsection{Vision-Language Instruction Tuning}
Inspired by the recent success of instruction tuning on LLMs~\cite{instructgpt, selfinstruct, benchmarking}, researchers have adopted vision-language instruction tuning to improve instruction following capabilities of Multimodal LLMs~\cite{Zhu2023MiniGPT4EV, Luo2023ValleyVA, Dai2023InstructBLIPTG, Li2023VideoChatCV, Zhang2023VideoLLaMAAI, Maaz2023VideoChatGPTTD}. 
This primarily entails producing high-quality data with human instructions, which can be categorized into two technical branches.
The first branch~\cite{Xu2022MultiInstructIM, Dai2023InstructBLIPTG, Li2023M3ITAL} integrates available multi-modal benchmark datasets and converts them to instruction format, with efforts like MultiInstruct~\cite{Xu2022MultiInstructIM}, InstructBLIP~\cite{Dai2023InstructBLIPTG}, and $\text{M}^3\text{IT}$~\cite{Li2023M3ITAL}. 
The second branch~\cite{Zhu2023MiniGPT4EV, Liu2023VisualIT, Li2023MIMICITMI, Li2023VideoChatCV, Luo2023ValleyVA} leverages LLMs such as ChatGPT~\cite{chatgpt} and GPT-4~\cite{OpenAI2023GPT4TR} to create more diverse dialog-style data. Approaches like MiniGPT4~\cite{Zhu2023MiniGPT4EV}, LLaVA~\cite{Liu2023VisualIT}, MIMIC-IT~\cite{Li2023MIMICITMI}, VideoChat~\cite{Li2023VideoChatCV}, and Valley~\cite{Luo2023ValleyVA} obtain detailed visual descriptions and build image-centric or video-centric conversation data from LLMs. 
However, they all neglect the time-aware user requests for video understanding. To address this, we propose a time-aware instruction tuning dataset to enhance the time-vision association ability of Multimodal LLMs.

\subsection{Video Temporal Localization}
Temporal localization is a foundational capability in video understanding tasks, particularly for untrimmed long videos. There have been miscellaneous time-sensitive video tasks, including temporal video grounding~\cite{Wang2021NegativeSM,Luo2023TowardsGV}, dense video captioning~\cite{zhu2022end, wang2021end, Yang2023Vid2SeqLP}, video summarization~\cite{zhang2016videosumm,Apostolidis2021VideoSU,Song2015TVSumSW,Gygli2014CreatingSF}, video highlight detection~\cite{Lei2021QVHighlightsDM, Moon2023QueryD}, and step localization~\cite{shen2021learning, dvornik2023stepformer, Zala2023HierarchicalVR}, etc. These tasks necessitate explicit associations between video semantics and the corresponding timestamps. Previous studies~\cite{Wang2021NegativeSM,wang2021end,zhang2016videosumm,dvornik2023stepformer} tend to settle each task separately on specialized downstream datasets. Although recent works~\cite{zhang2022unifying, Jiang2022JointVS, Lei2021QVHighlightsDM, lin2023univtg, liu2022umt} make preliminary attempts to bridge several tasks, a generalist paradigm based on LLMs is under exploration.
In this paper, we unify a wide range of time-sensitive video tasks in language modeling format and take a first step to leverage LLMs. 

\section{Method}
In this section, we present \modelname, a VidLLM featuring two novel modules: a timestamp-aware frame encoder and a sliding video Q-Former. These modules enhance our \modelname's ability to localize temporally and understand long videos ($\S$~\ref{subsec:arch}). 
To further empower \modelname to follow human instructions across time-sensitive video tasks, we collect an instruction-tuning dataset named \datasetname ($\S$~\ref{subsec:it}). This dataset comprises 6 tasks and 125K instances. Based on \datasetname, we perform instruction tuning on our model to unlock its full potential.

\subsection{\modelname Architecture}
\label{subsec:arch}

\subsubsection{Overview}
\modelname is composed of a time-aware frame encoder, a sliding video Q-Former, and a large language model, as depicted in Fig.~\ref{fig:arch}. Given an input video, the frame encoder first extracts visual and timestamp features for each frame independently. Next, the video Q-Former models temporal relations across frames within a sliding window to produce video tokens. Finally, these video tokens are concatenated with optional transcribed speech and user instructions, which are then fed into the LLM to generate responses. 

\vspace{-8pt}
\subsubsection{Timestamp-aware Frame Encoder}
Previous studies typically separate the modeling of visual semantics and their respective timestamp information of input frames~\cite{Li2023VideoChatCV, Zhang2023VideoLLaMAAI}. 
For example, VideoChat~\cite{Li2023VideoChatCV} utilizes a visual encoder to process visual frame semantics but an LLM to receive timestamp information, e.g. ``\texttt{This video contains 8 frames sampled at 2s, 4s,}  $\dots$\texttt{, 16s}''. 
As a result, this method fails to directly capture the time when a visual event occurs. Some alternative approaches add learnable position (time) embeddings to the visual tokens~\citep{Zhang2023VideoLLaMAAI}. However, this only enables models to discern frame order, lacking precision in determining the exact temporal moment. 

To mitigate these issues, we introduce a time-aware frame encoder (green block in ~\cref{fig:arch}) inspired by InstructBLIP~\citep{Dai2023InstructBLIPTG}. 
Given a video input $\mathbf{V}\in \mathbb{R}^{T\times H\times W\times 3}$,
the encoder first uses a pre-trained image encoder, i.e. ViT~\citep{Radford2021LearningTV, Sun2023EVACLIPIT}, to encode each frame and obtain frame features. Subsequently, an image Q-Former further compresses the frame tokens. 
As \cref{fig:arch} illustrates, the Q-Former takes $N_I$ learnable queries in dimension $D_Q$ as input. These queries interact with the frame features via cross-attention and update the initial queries to final $N_I$ visual tokens in dimension $D_Q$ ~\citep{Li2023BLIP2BL, Dai2023InstructBLIPTG}. 
Importantly, during visual token extraction, we add the frame's timestamp\footnote{Vid2Seq~\citep{Yang2023Vid2SeqLP} shows comparable performance using relative and absolute timestamps (Tab.13, N=500). For chat-friendliness and simplicity, we opt for the absolute ones.}, e.g., ``\texttt{This frame is sampled at 2s.}'', as a condition to the Q-Former to fuse the visual and timestamp information. 
While our approach shares structural similarities with InstructBLIP's Q-Former, the motivations differ. InstructBLIP uses instructions (e.g., ``\texttt{Choose the correct option to the following question:} $\dots$'' for VQA tasks) as additional input to extract task-relevant visual tokens, whereas our Q-Former takes timestamp descriptions to bind time information to the visual tokens.

\vspace{-5pt}
\subsubsection{Sliding Video Q-Former}
\label{subsubsec:sliding-qformer}
After applying the time-aware frame encoder, we obtain $T\times N_I$ visual tokens for a $T$-frames video input. 
Since frames are encoded independently, the temporal relationship across frames has not been modeled yet. To this end, we 
incorporate a sliding video Q-Former (yellow block in Fig.~\ref{fig:arch}) to enhance the feature fusion in the temporal dimension.
The video Q-Former mirrors the structure of the image Q-Former, it takes $N_V$ learnable queries in dimension $D_Q$ as input without timestamps.  
We design a sliding window of length $L_{W}$ 
 and within each window utilizing the video Q-Former extract $N_V$ video tokens 
from $L_{W}$ frames.
By sliding the video Q-Former in strides of $S$, we can represent the input video as $(T/S) \times N_V$ video tokens.

Considering the 3D nature of videos and the redundancy in space-time information, the original sequence of visual tokens (i.e., patches in all frames) can be extremely long~\cite{Ren2023TESTA}. Thus, it's crucial to condense video information to a reduced number of video tokens, thereby decreasing the computation burden on the LLM. 
However, previous work~\citep{Li2023VideoChatCV, Zhang2023VideoLLaMAAI, jin2023chatunivi, Liu2023VideoTellerEC} 
usually set a fixed number of video tokens $N_V$, such as 32, which can result in severe visual semantics degradation
 when the number of input frames $T$ is large. Concretely, 
we define the compression rate $R$ as the ratio of the number of original visual tokens to the number of final video tokens.
The compression rate for previous work like Video-LLaMA~\citep{Zhang2023VideoLLaMAAI} is:
\begin{equation}
\small
    R=(T \times N_P) / N_V,
\end{equation}
where $N_P$ is the number of patches of each frame. This ratio increases with the number of input frames $T$ and can cause excessive compression for long videos. 
With our sliding video Q-Former, our compression rate $R^{'}$ becomes a constant value:
\begin{equation}
\small
    R^{'} = \frac{T \times N_P}{(T/S) \times N_V} = \frac{S\times N_P}{N_V},
\end{equation}
retaining richer semantics for long videos. By adjusting the stride $S$, we can control the final number of video tokens according to the computation budget. 
Finally, we use a linear layer to transform the dimension $D_Q$ of video tokens to match the dimension $D_{LLM}$ of the LLM embedding space.

\subsubsection{Large Language Model}
Ultimately, we concatenate inputs from various modalities, e.g., the video tokens $\mathbf{X}_v$, text query tokens $\mathbf{X}_{q}$ (including optional transcribed speech and user instruction), and feed these into a large language model to generate reasonable and coherent responses (answers) $\mathbf{X}_{a}$. Here, $\mathbf{X}_v$, $\mathbf{X}_{q}$, and $\mathbf{X}_{a}$ have the same token embedding dimension $D_{LLM}$. 
The training of the VidLLM typically utilizes a two-stage training framework. The first stage pre-trains the model using large-scale image/video-text pairs for vision-language alignment~\citep{Li2023BLIP2BL, Xu2021VideoCLIPCP, Liu2023FETVAB, Li2023VITATECSAD, ren-etal-2023-delving, Ren2021LearningRA}. The second stage finetunes the model with instruction data for instruction following. 
Considering computing efficiency, we reuse the checkpoints of the existing open source models after the first stage training (see $\S$~\ref{subsec:implement}), conducting only instruction tuning. 
During the training procedure, we utilize the language modeling loss for generating target answers $\mathbf{X}_{a}$ with length $L_{T}$, which serves as the objective function:
\begin{equation}
\small
\begin{split}
    \mathcal{L} &= - \log P_\theta (\mathbf{X}_a \mid \mathbf{X}_v, \mathbf{X}_q) \\
    &= -\sum_{i=1}^{L_T} \log P_\theta \left(x_i \mid \mathbf{X}_v, \mathbf{X}_{q}, \mathbf{X}_{a, <i} \right),
\end{split}
\end{equation}
where $\theta$ is the trainable parameters, and $\mathbf{X}_{a, <i}$ refers to the answer tokens preceding the current prediction token $x_i$. 
To better adapt the LLM to video tasks, we apply the parameter-efficient fine-tuning method, LoRA~\citep{Hu2021LoRALA}. 

\subsection{Instruction Data \datasetname}
\label{subsec:it}

\begin{figure}[t]
    \centering
    \includegraphics[width=\linewidth]{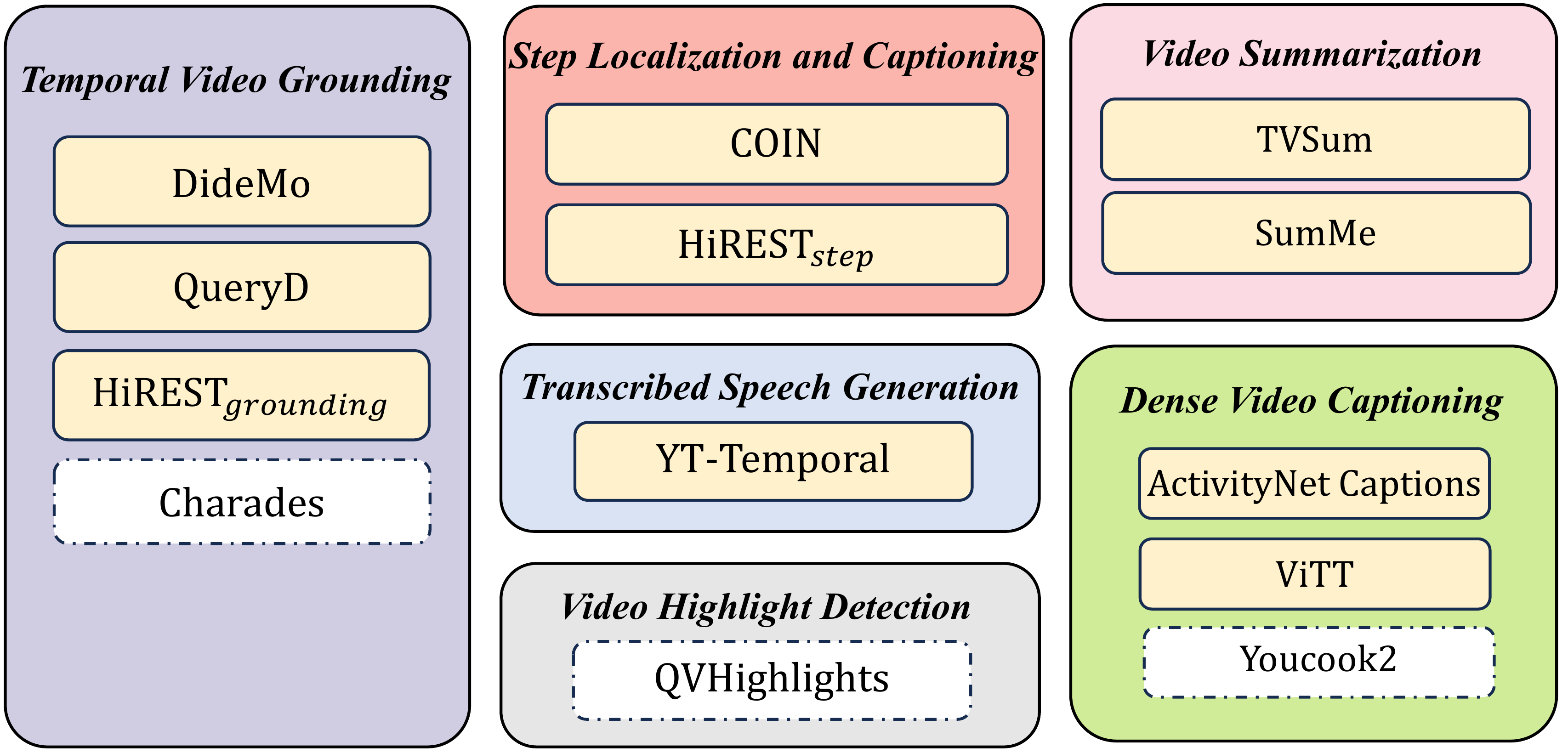}
    \caption{Involved tasks and datasets in the time-aware instruction tuning dataset. The held-in datasets are colored with yellow blocks while the held-out datasets are in dashed white blocks. }
    \label{fig:dataset}
\end{figure}

To boost \modelname's ability to understand time-sensitive human instructions, we introduce \datasetname, a video-centric instruction-tuning dataset involving timestamps. This dataset integrates a wide range of timestamp-associated video datasets and is characterized by long-form videos.

\subsubsection{Task Coverage}
\datasetname encompasses 6 longstanding timestamp-related video tasks, i.e., (1) dense video captioning, (2) temporal video grounding, (3) step localization and captioning, (4) video summarization, (5) video highlight detection, as well as (6) transcribed speech generation. It also incorporates 12 specific datasets~\citep{Hendricks2017LocalizingMI, Oncescu2020QUERYDAV, Zala2023HierarchicalVR, Gao2017TALLTA, Krishna2017DenseCaptioningEI, Huang2020MultimodalPF, Zhou2017TowardsAL, Song2015TVSumSW, Gygli2014CreatingSF, Lei2021QVHighlightsDM, Tang2019COINAL, Zellers2022MERLOTRN} derived from different domains as illustrated in \cref{fig:dataset}. Please refer to Appendix~\ref{sec:task-cov} for details.
Our dataset accommodates prevalent user requests involving video timestamps when interacting with AI assistants in real-world applications.

\vspace{-5pt}
\subsubsection{Data Construction}
We convert the above datasets into an instruction-following format to obtain high-quality video-centric instruction data. The construction process comprises two main steps including (1) instruction writing and (2) answer formatting.

\vspace{-10pt}
\paragraph{Step I: Instruction Writing. } The quality and diversity of instructions are essential in the construction process. We manually write well-designed instructions for each task as a good starting. Then we utilize GPT-4~\citep{OpenAI2023GPT4TR} to extend more diverse and flexible expressions based on the manual initialization. Eventually, we manually select and refine the LLM-generated instructions to obtain the final version. Inspired by the observation in $\text{M}^3\text{IT}$~\cite{Li2023M3ITAL} that using around five instructions per task is sufficient, we generate six high-quality instructions for each task. Specific instructions designed for each task are depicted in Appendix~\ref{sec:ins4task}.

\vspace{-10pt}
\paragraph{Step II: Answer Formatting. } Based on the written instructions, we further reformulate the task outputs into a user-friendly natural language response paradigm (format details are provided in Appendix~\ref{sec:ins4task}). 
Considering the involved video datasets are manually collected, the overall quality of \datasetname data is guaranteed. 

\cref{table:datastat} compares our \datasetname data with existing video-centric instruction tuning data, revealing our significant advantages across data scale, task diversity, and video length. 
Appendix~\ref{sec:contribution-analysis} provides contribution analysis of each task to model performance. Overall, all 6 tasks are beneficial. 


\begin{table}[t!]
\centering
\begin{adjustbox}{max width=\linewidth}
\begin{tabular}{@{}lcccc@{}}
\toprule
Dataset &  Time-aware & \# Tasks   & \# Samples  & Avg. video len \\ \midrule
VideoChat~\cite{Li2023VideoChatCV} & \xmark & 3   &  11K   &  18.0s  \\ 
Valley~\cite{Luo2023ValleyVA} & \xmark & 2 & 73K & 21.9s \\
Video-ChatGPT~\cite{Maaz2023VideoChatGPTTD} & \xmark & 1   &  100K   &  115.2s  \\ 
\datasetname (Ours) & \cmark & 6 & 125K &  190.8s \\
\bottomrule
\end{tabular}
\end{adjustbox}
\caption{Comparison with existing video instruction tuning data.}
\label{table:datastat}
\end{table}


\begin{table*}[t!]
\centering
\begin{adjustbox}{max width=\linewidth}

\begin{tabular}{@{}lclll|ll|ll@{}}
\toprule
\multirow{3}{*}{Model} & \multirow{3}{*}{\begin{tabular}[c]{c}LLM\\Size\end{tabular}} & \multicolumn{3}{c|}{Dense Video Captioning} & \multicolumn{2}{c|}{Highlight Detection} & \multicolumn{2}{c}{Temporal Grounding} \\
                 &  & \multicolumn{3}{c|}{YouCook2~\citep{Zhou2017TowardsAL}}                 & \multicolumn{2}{c|}{QVHighlights~\citep{Lei2021QVHighlightsDM}}               & \multicolumn{2}{c}{Charades-STA~\citep{Gao2017TALLTA}}              \\ \cmidrule(l){3-9} 
                 &  & SODA\_c   & CIDEr        & F1 Score          & mAP                  & HIT@1        & 
                 R@1 {\footnotesize (IoU=0.5)}          
                 & 
                 R@1 {\footnotesize (IoU=0.7)} 
                 \\ \midrule
\rowcolor{dt!50}
\multicolumn{9}{c}{Multi-model Pipelines} \\
InstructBLIP~\citep{Dai2023InstructBLIPTG}+ChatGPT~\citep{chatgpt} & -  &   0.5 &  1.0  & 8.2                &  25.6                    &  53.0 &      7.0                 &       1.0                      \\
VideoChat-Text {\footnotesize (w/ ChatGPT)}~\citep{Li2023VideoChatCV} & -    &    0.4          &  0.9               &       8.4                  &    17.5                  &     31.0     &  9.0                     &     3.0                   \\ \midrule
\rowcolor{dt!50}
\multicolumn{9}{c}{End2end VidLLMs} \\
\demph{Video-LLaMA}~\citep{Zhang2023VideoLLaMAAI} & \demph{13B}   &   \demph{0.0}   &  \demph{0.0}   &   \demph{0.1}  &   \demph{11.1}   & \demph{15.6}  &   \demph{2.1}     &   \demph{0.6} \\
\demph{VideoChat-Embed}~\citep{Li2023VideoChatCV} & \demph{13B}   &   \demph{0.5}   &  \demph{1.0}   &   \demph{7.0}  &   \demph{14.2}     &   \demph{18.9}  &   \demph{2.6}   & \demph{0.8} \\
Valley~\citep{Luo2023ValleyVA} & 7B  &  0.1 &   0.0  & 1.5  &  10.9 &   15.2 &  4.7 &    1.6 \\
Video-LLaMA~\citep{Zhang2023VideoLLaMAAI} & 7B & 0.0 & 0.0 & 0.1 & 11.3 & 15.6 & 2.7 & 1.2 \\
VideoChat-Embed~\citep{Li2023VideoChatCV} & 7B      &    0.2   &  0.6   &   3.4   &    13.1    &   18.1  &  3.2          &   1.4   \\
\modelname (Ours)  & 7B    &   $\textbf{1.2}_\impro{1.0}$ &  $\textbf{3.4}_\impro{2.8}$  & $\textbf{12.6}_\impro{9.2}$  &   $\textbf{14.5}_\impro{1.4}$  &  $\textbf{23.9}_\impro{5.8}$  &   $\textbf{32.2}_\impro{27.5}$   &   $\textbf{13.4}_\impro{11.8}$  \\ \bottomrule
\end{tabular}

\end{adjustbox}
\caption{Zero-shot performance on three video tasks of dense captioning (YouCook2), highlight detection (QVHighlights), and temporal grounding (Charades). For all metrics, higher is better. We gray out methods that use 13B LLMs for fair comparison. The performance gain in {\color[HTML]{32CB00}green} is compared to the best-performing 7B model. We do not use transcribed speech for evaluation.}
\label{table:zeroshot}
\end{table*}

\section{Experiments}
\subsection{Implementation Details}
\label{subsec:implement}
We take ViT-G/14 from EVA-CLIP~\citep{Sun2023EVACLIPIT} as the image encoder and LLaMA-2 (7B)~\citep{touvron2023llama2} as the language foundation model. The parameters of the image Q-Former are initialized from InstructBLIP's checkpoint, while the video Q-Former is initialized from Video-LLaMA's checkpoint. 
We finetune our \modelname on \datasetname and Valley~\citep{Luo2023ValleyVA} for 3 epochs, using a batch size of 32, with a single 8-V100 (32G) machine. As shown in Fig.~\ref{fig:arch}, the parameters of ViT and LLM are frozen, while those of image Q-Former, video Q-Former, and linear layer are tuned. The rank in LoRA is 32. 
The window size $L_W$, stride $S$, and the number of video tokens $N_V$ per window are 32. The number of input frames is 96. Please refer to Appendix~\ref{sec:hyper-parameters} for additional hyper-parameters. 


\subsection{Evaluation Setups}
\paragraph{Tasks, Datasets and Evaluation Metrics.}
We evaluate our model on three tasks of long video understanding, i.e., dense captioning, temporal grounding, and highlight detection, in a zero-shot setting. The evaluation datasets include YouCook2~\citep{Zhou2017TowardsAL}, Charades-STA~\citep{Gao2017TALLTA}, and QVHighlights~\citep{Lei2021QVHighlightsDM}. 
See Appendix~\ref{sec:evaluation} for details of evaluation metrics. 
\vspace{-10pt}
\paragraph{Heuristic Rules for Parsing LLM Outputs.} It is important to note that the outputs generated by LLMs may include colloquial expressions, leading to a wide range of response variations. Accordingly, we carefully devise a considerable number of heuristic rules to guarantee that predicted answers can be accurately extracted from the model's responses for the computation of final metrics. 

\vspace{-8pt}
\paragraph{Compared Methods.}
We compare our model with two branches of baselines. 
\textbf{(1) Multi-model Pipielines}, including VideoChat-Text~\citep{Li2023VideoChatCV}, InstructBLIP~\citep{Dai2023InstructBLIPTG}+ChatGPT~\citep{chatgpt}. These pipelines integrate specialized visual models with ChatGPT, which firstly convert video semantics (e.g. frame descriptions, clip captions or action tags) into textual descriptions and then leverage ChatGPT to process all inputs to solve the target task. See Appendix~\ref{sec:pipeline} for more details. 
\textbf{(2) End2end Models}, including Valley~\citep{Luo2023ValleyVA}, VideoChat-Embed~\citep{Li2023VideoChatCV}, Video-LLaMA~\citep{Zhang2023VideoLLaMAAI} with 7B LLMs. These models directly take videos as inputs and generate responses in an end2end manner. 

\begin{table}[t!]
\centering
\begin{adjustbox}{max width=\linewidth}

\begin{tabular}{@{}llll@{}}
\toprule
Model &  SODA\_c   & CIDEr  & F1 Score \\ \midrule
\modelname & 3.1 & 10.3 & 19.5 \\ 
~~w/o sliding video Q-Former & $2.1_\degra{1.0}$ & $7.5_\degra{2.8}$ & $16.5_\degra{3.0}$ \\
~~w/o timestamp-aware frame encoder & $1.8_\degra{0.3}$ & $6.1_\degra{1.4}$ & $14.2_\degra{2.3}$ \\
\bottomrule
\end{tabular}
\end{adjustbox}
\caption{Ablation study. We remove the two key modules respectively, then finetune and evaluate our model on YouCook2.}
\vspace{-8pt}
\label{table:ablation}
\end{table}

\subsection{Zero-shot performance}
Tab.~\ref{table:zeroshot} shows the zero-shot performance of \modelname (7B), which outperforms previous VidLLMs (7B/13B) in all tasks. 

\vspace{-10pt}
\paragraph{Dense Video Captioning.}
This task on YouCook2 is quite challenging.  The model is required to accurately identify roughly 8 essential cooking steps within the average video duration of 320 seconds, alongside providing faithful descriptions that match the visual content. Moreover, the specialized nature of cooking amplifies the task complexity, thereby challenging the model's generalizability. 
Existing end-to-end VidLLMs struggle with precise moment localization, as evidenced by the low F1 score of 3.4 achieved by the top-performing VideoChat-Embed model. Such imprecision in moment localization significantly impacts the captioning evaluation, with both SODA\_c and CIDEr metrics approaching zero. 
Compared to them, our model achieves remarkable performance gains exceeding the previous SOTA by +1.0 SODA\_c, +2.8 CIDEr, and +9.2 F1 score. This reveals that \modelname effectively processes lengthy videos with precise temporal localization capability. 
Moreover, our performance also significantly surpasses the multi-model pipelines powered by ChatGPT (F1 score from 8.4 to 12.6), which demonstrates both the challenging nature of this task and the superiority of our model in processing long videos.


\begin{table}[t!]
\centering
\begin{adjustbox}{max width=.8\linewidth}

\begin{tabular}{@{}lccc@{}}
\toprule
Model &  SODA\_c   & CIDEr  & F1 Score \\ \midrule
Video-LLaMA~\citep{Zhang2023VideoLLaMAAI} &  1.8  &  6.1  & 14.2  \\
VideoChat-Embed~\citep{Li2023VideoChatCV}   &  2.2    &  8.1   &  15.6 \\
\modelname (Ours) &  \textbf{3.1}  &  \textbf{10.3}  &  \textbf{19.5} \\ 
\bottomrule
\end{tabular}
\end{adjustbox}
\caption{Performance of various VidLLMs (7B) after fine-tuneing on YouCook2 (w/o using the \datasetname dataset).}
\vspace{-8pt}
\label{table:ft}
\end{table}

\vspace{-8pt}
\paragraph{Highlight Detection.}
While the dense video captioning task focuses on localizing events at the clip level, this task requires a more fine-grained video comprehension at the frame level. 
For an input video, it aims to output the times and the related salient scores of highlight frames.
Overall, 
our model achieves a 14.5 mAP and 23.9 HIT@1 on QVHighlights, 
surpassing the previous VidLLMs by +1.4 and +5.8 points, respectively. 
This highlights the contribution of our timestamp-aware frame encoder in identifying the salient semantics of each frame. 
Moreover, this task is a held-out task in \datasetname, indicating the generalization ability of our model on novel tasks. 
As for multi-model pipeline approaches, they achieve even stronger performance. We speculate that this is due to the format of highlight detection being more compatible with their methods,
as the model receives a series of joint timestamp-visual descriptions for input frames. 
This enables the frame-by-frame assessment by the LLM, facilitating more accurate judgments. 

\vspace{-8pt}
\paragraph{Temporal Grounding.} 
This task aims to identify the corresponding timestamp described by a query sentence.
\modelname achieves 32.2 points on ``R@1, IoU=0.5'' of the Charades-STA dataset, which surpasses the previous SOTA end2end VidLLM, i.e. Valley, by a substantial margin (+27.5).
This demonstrates that our model excels at accurately localizing the video moment content referred to a given text query. 
Notably, \modelname gains the most improvements on the temporal grounding task, we argue that this task mainly emphasizes the temporal localization capability of long videos which is exactly the best advantage of \modelname.



\begin{figure}[t]
\centering
\includegraphics[width=\linewidth]{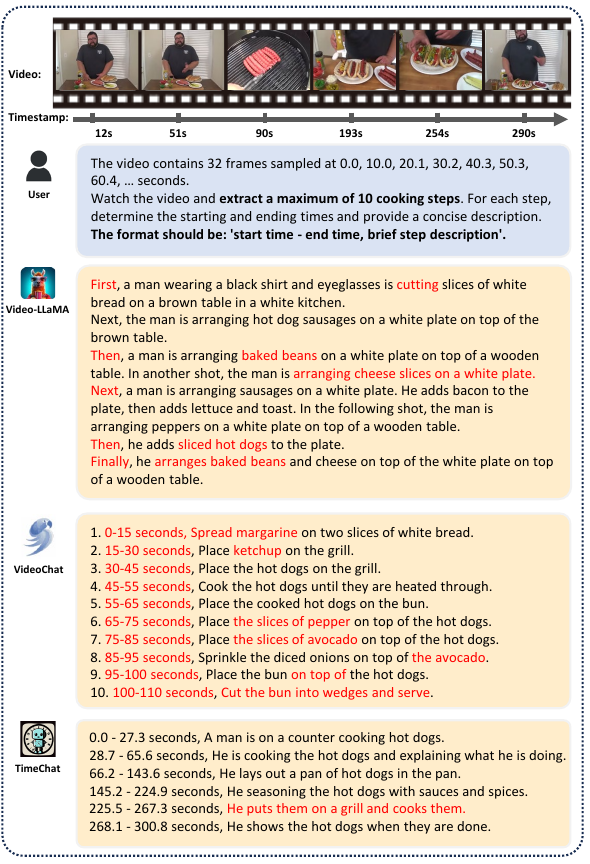}
\caption{Case study (in zero-shot). The red text highlights inaccuracies in the model-generated timestamp or description.}
\label{fig:case}
\end{figure}


\subsection{Qualitative Evaluation}
\paragraph{Compared to Existing VidLLMs.}
Fig.~\ref{fig:case} presents qualitative comparisons between \modelname and other VidLLMs in zero-shot settings. 
Video-LLaMA falls short in fully adhering to the user instruction, as it only describes the cooking steps without the corresponding start and end timestamps for each step. VideoChat, on the other hand, produces captions that fit the requested format but misplaces the timing of all the steps. Despite this, the generated description from VideoChat includes several hallucinations~\citep{Li2023EvaluatingOH}, such as references to ``pepper'' and ``avocados'' that are not present in the video. In contrast, \modelname demonstrates improved temporal localization and summarization capabilities compared to the previous models. 
It successfully matches the content of the video for almost all extracted clips. 
Furthermore, the occurrence of hallucinations is significantly reduced. However, there is still room for improvement in terms of enhancing the richness and details in the summarization generated by our model.

\vspace{-8pt}
\paragraph{Generalized to New Domains.}
In Appendix~\ref{sec:domain-generalize}, we show qualitative results in new domains such as movie~\cite{Yue2023Movie101AN} and egocentric videos~\cite{Grauman2021Ego4DAT}, demonstrating the generalization of \modelname to novel scenarios. This generalization is a key characteristic towards a practical video assistant and represents a fundamental difference between LLM-based \modelname and the current specialized models tailored for specific downstream datasets.
More cases can be found in Appendix~\ref{sec:more-res}.

\subsection{Ablation Study}

\begin{table*}[t!]
\centering
\begin{adjustbox}{max width=\linewidth}

\begin{tabular}{@{}lccccc|cc|cc@{}}
\toprule
\multirow{3}{*}{Model} & \multirow{3}{*}{\begin{tabular}[c]{c}Generalist\\Model\end{tabular}} & \multirow{3}{*}{\begin{tabular}[c]{c}Finetune\\Epochs\end{tabular}} & \multicolumn{3}{c|}{Dense Video Captioning} & \multicolumn{2}{c|}{Highlight Detection} & \multicolumn{2}{c}{Temporal Grounding} \\
             &    &      & \multicolumn{3}{c|}{YouCook2~\citep{Zhou2017TowardsAL}}                 & \multicolumn{2}{c|}{QVHighlights~\citep{Lei2021QVHighlightsDM}}               & \multicolumn{2}{c}{Charades-STA~\citep{Gao2017TALLTA}}              \\ \cmidrule(l){4-10} 
             &     &     & SODA\_c   & CIDEr        & F1 Score          & mAP                  & HIT@1        & R@1 {\footnotesize (IoU=0.5)}          & R@1 {\footnotesize (IoU=0.7)}                  \\ \midrule
\demph{Vid2Seq}~\cite{Yang2023Vid2SeqLP}   &  \xmark  &  \demph{40} & \demph{7.9} &  \demph{47.1} &  \demph{27.3}  &  \demph{-}          &   \demph{-}         &   \demph{-}        &     \demph{-}        \\
\demph{QD-DETR}~\citep{Moon2023QueryD} &  \xmark & \demph{200} & \demph{-}  & \demph{-} & \demph{-}         &  \demph{38.9}   &   \demph{62.4}  & \demph{-}&   \demph{-} \\
\demph{${\text{QD-DETR}}_{w/ Audio}$}~\citep{Moon2023QueryD} &  \xmark & \demph{200} & \demph{-}  & \demph{-} & \demph{-}         &  \demph{39.0}    &   \demph{62.9}  & \demph{-} &   \demph{-}         \\
\demph{MMN}~\citep{Wang2021NegativeSM}  &  \xmark & \demph{18} & \demph{-}  & \demph{-} & \demph{-}   &   \demph{-}         &    \demph{-}  & \demph{50.5} &   \demph{29.7}                  \\ 
\demph{VDI}~\citep{Luo2023TowardsGV}  &  \xmark & \demph{18} & \demph{-}  & \demph{-} & \demph{-}    &   \demph{-}         &    \demph{-}  & \demph{52.3}  &   \demph{31.4}    \\  \midrule
\modelname (Ours)  &  \cmark & 3 &  3.4  &  11.0  & 19.5 &   21.7 &  37.9  & 46.7  &  23.7 \\ \bottomrule
\end{tabular}

\end{adjustbox}
\caption{Supervised performance compared to task-specific models. They use specific objective functions and extra fine-tuning epochs to better fit the downstream dataset. In contrast, \modelname exhibits a broad versatility 
 and generalization across various tasks and domains.}
\label{table:supervised}
\end{table*}

We conducted an ablation study based on YouCook2 to assess the efficacy of key designs in our \modelname. 
As illustrated in Tab.~\ref{table:ablation}, 
when the sliding video Q-Former is removed, the number of final visual tokens decreases from 96 to 32, resulting in a $3\times$ information compression rate. This reduction in semantic information leads to a decrease in the alignment between the generated descriptions and the video content. Specifically, the SODA\_c 
metric decreases by 1.0, while the CIDEr metric decreases by 2.8. Additionally, the accuracy of timestamps (measured by F1 score) decreases by 3.0. 
In the case of the removal of the timestamp-aware frame encoder, the model's ability to temporally ground the descriptions diminishes dramatically, as indicated by a decrease of 2.3 in the F1 score. 
These results highlight the effectiveness of the two novel modules in the model.

\subsection{Further Analysis}
\vspace{-5pt}
We provide further analysis to validate the superiority of our model. 
To demonstrate that the performance gain of our model is not solely attributed to the new \datasetname dataset, but also to the improvements in our model architecture, we conduct fine-tuning and evaluation using only the YouCook2 dataset. In this setup, we initialize our model with existing open-source checkpoints (see $\S$~\ref{subsec:implement}). For all the models, we finetune their Q-Formers and apply LoRA~\citep{Hu2021LoRALA} for their LLMs. 
Tab.~\ref{table:ft} presents the results, showing that our model consistently outperforms previous models across all metrics, with increases of +2.2 CIDEr and +3.9 F1 score.

\begin{figure}[t]
\centering
\includegraphics[width=\linewidth]{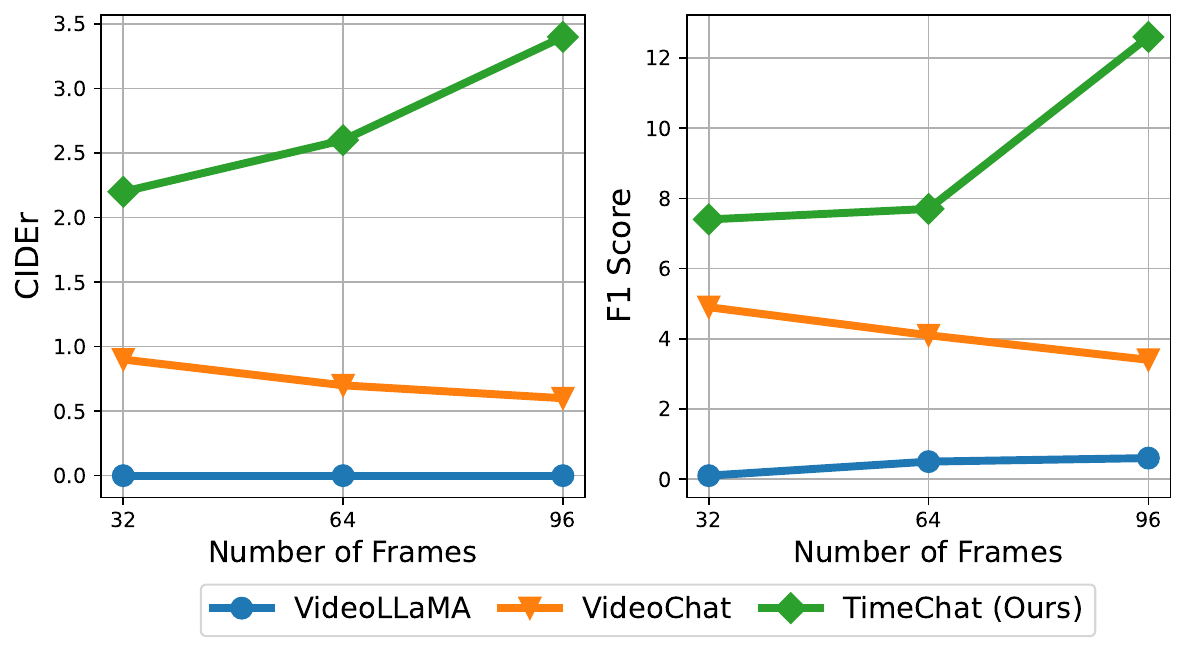}
\caption{Comparison of CIDEr and F1 score w.r.t the number of input frames on the YouCook2 dataset (zero-shot).}
\label{fig:vary-frames}
\end{figure}

In ~\cref{fig:vary-frames}, we examine the performance scalability of our model with respect to the number of input frames~\citep{Ren2023TESTA}. As mentioned in $\S$~\ref{subsubsec:sliding-qformer}, previous models like Video-LLaMA and VideoChat compress excessive information for long videos, resulting in minimal performance impact when increasing the number of input frames from 32 to 96.  
In contrast, our \modelname decouples the number of frames $T$ and the compression rate $R^{'}$ using the sliding video Q-Former. Our curve exhibits linear improvement in performance as the number of frames increases, showcasing superior scalability.

\subsection{Comparison with Specialized Models}

In this subsection, we compare our generalist model, \modelname, with state-of-the-art specialized models on the three tasks, respectively. Given that all specialized models have been fine-tuned on specific datasets, we also finetune our model for a fair comparison. 
As shown in Tab.~\ref{table:supervised}, after fine-tuning, \modelname has made further performance gains, e.g., +6.9 F1 score on YouCook2, +16.9 HIT@1 on QVHighlights, and +16.4 R@1 (IoU=0.5) on Charades-STA. 
Nonetheless, there is still much room for boosting our approach compared to specialized models, whose superior performance arises from task-specific designs.
For example, Vid2Seq~\cite{Yang2023Vid2SeqLP} pretrains on YT-Temporal-1B~\citep{Zellers2022MERLOTRN}, which contains much more high-quality long videos than the tuning dataset we used. QD-DETR~\citep{Moon2023QueryD} employs a special saliency token for saliency prediction and introduces 4 loss functions for training, while our model trains purely through language modeling. 
In addition, these models also used much more fine-tuning steps to better fit the downstream dataset. 
However, as a generalized model, \modelname exhibits a strong generalization ability in zero-shot scenarios, multi-task, and multi-domain settings, which is not present in those expert models. 
Achieving state-of-the-art performance on every task is not the major goal of this paper, and we leave this as future work.

\section{Discussion and Conclusion}
We present \modelname, a time-sensitive VidLLM for long video understanding. 
Benefiting from the novel time-aware frame encoder, sliding video Q-Former, and instruction tuning on \datasetname, our model demonstrates strong temporal localization capabilities that were absent in previous VidLLMs. 
Through its ability to identify significant events within lengthy videos, pinpoint events' start and end times, and generate concise summarization, \modelname makes a crucial step toward an intelligent video assistant. In the future, we will make architectural advances to improve video semantic density while reducing spatial-temporal redundancy. We will also collect more diverse and high-quality instruction-tuning data to broaden the time-related applications.

\section*{Acknowledgements}
This work is supported in part by a Huawei Research Grant and National Natural Science Foundation of China (No. 62176002). Xu Sun is the corresponding author. 

{
    \small
    \bibliographystyle{ieeenat_fullname}
    \bibliography{main}
}


\clearpage

\clearpage
\appendix

\section{Task Coverage in \datasetname}
\label{sec:task-cov}
\datasetname encompasses 6 longstanding timestamp-related video tasks and incorporates 12 specific datasets derived from different domains. 

\vspace{-5pt}
\paragraph{Dense Video Captioning (DVC).} This task unifies the event localization and event captioning subtasks. It detects a series of events in the given video and outputs the corresponding timestamps and descriptions. We gather ActivityNet Captions~\cite{Krishna2017DenseCaptioningEI}, ViTT~\cite{Huang2020MultimodalPF}, and YouCook2~\cite{Zhou2017TowardsAL} datasets to facilitate the narration of significant events for users when watching long videos.

\vspace{-5pt}
\paragraph{Temporal Video Grounding (TVG).} This task aims to predict a timestamp boundary including the start and end time in the video given a natural language query. We include DiDeMo~\cite{Hendricks2017LocalizingMI}, QuerYD~\cite{Oncescu2020QUERYDAV}, $\text{HiREST}_{grounding}$~\cite{Zala2023HierarchicalVR}, and Charades-STA~\cite{Gao2017TALLTA} datasets to achieve accurate moment localization when users interact with natural language.

\vspace{-5pt}
\paragraph{Step Localization and Captioning (SLC).} This task is designed to automatically segment and describe significant steps in a long untrimmed video, which is useful for instructional videos. We incorporate two datasets including COIN~\cite{Tang2019COINAL} and $\text{HiREST}_{step}$ to fulfill key steps detecting when processing noisy instructional videos under the cooking, repairing, or assembling furniture scenarios.

\vspace{-5pt}
\paragraph{Video Summarization (VS).} The goal is to create a compressed set of frames or clip shots to represent the most informative content of the given video. TVSum~\cite{Song2015TVSumSW} and SumMe~\cite{Gygli2014CreatingSF} datasets are compiled to achieve an efficient video overview for busy stakeholders to save time.

\vspace{-5pt}
\paragraph{Video Highlight Detection (VHD).} Different from the video summarization, it identifies the most exciting, impressive, or emotional moments that may not cover the full scope of the original video. QVHighlights~\cite{Lei2021QVHighlightsDM} dataset is utilized to evaluate the highlight moment recommendation ability of AI assistants.

\vspace{-5pt}
\paragraph{Transcribed Speech Generation (TSG).} The objective of this task is to predict the speech content and its corresponding start and end timestamps based on visual signals in the video. This task can be regarded as a weakly-supervised event localization and description task. We use the YT-Temporal-1B dataset~\citep{Zellers2022MERLOTRN}. The original dataset includes 18 million narrated videos collected from YouTube, while we sample 31.6K videos from it for instruction tuning. Following Vid2Seq~\cite{Yang2023Vid2SeqLP}, we leverage Whisper-timestamped~\citep{lintoai2023whispertimestamped, radford2022robust} to automatically transcribe speech and use it as the target answer. 

Our dataset accommodates prevalent user requests involving video timestamps when interacting with AI assistants in real-world applications. 

\begin{table*}[t!]
\begin{adjustbox}{max width=\linewidth}
\centering
\begin{tabular}{l|p{15cm}}
\toprule
Key &  Value \\ \midrule
\rowcolor{dt!50}
\multicolumn{2}{c}{Dense Video Captioning} \\
Instruction Example &  Localize a series of activity events in the video, output the start and end timestamp for each event, and describe each event with sentences. \\ 
Output Format &  \colorbox{yyellow}{<timestamp\_start>} - \colorbox{yyellow}{<timestamp\_end>} seconds, \colorbox{ggreen}{<event\_description>}. $\cdots$ \\
Output Example & 90 - 102 seconds, spread margarine on two slices of white bread in the video. 114.0 - 127.0 seconds, place a slice of cheese on the bread. $\cdots$ \\ \midrule
\rowcolor{dt!50}
\multicolumn{2}{c}{Temporal Video Grounding} \\
Instruction Example &  Detect and report the start and end timestamps of the video segment that semantically matches the given textual query \colorbox{ppurple}{<query\_placeholder>}. \\ 
Output Format &  The given query happens in \colorbox{yyellow}{<timestamp\_start>} - \colorbox{yyellow}{<timestamp\_end>} seconds. \\ 
Output Example & The given query happens in 0.0 - 6.9 seconds. \\ \midrule
\rowcolor{dt!50}
\multicolumn{2}{c}{Step Localization and Captioning} \\
Instruction Example &  Identify and mark the video segments corresponding to a series of actions or steps, specifying the timestamps and describing the steps. \\ 
Output Format &  \colorbox{yyellow}{<timestamp\_start>} - \colorbox{yyellow}{<timestamp\_end>} seconds, \colorbox{ggreen}{<step\_description>}. $\cdots$ \\ 
Output Example & 21.0 - 22.0 seconds, begin to run up.  23.0 - 24.0 seconds, begin to jump up.  25.0 - 26.0 seconds, fall to the ground. \\ \midrule
\rowcolor{dt!50}
\multicolumn{2}{c}{Video Summarization} \\
Instruction Example & Generate a summarized version of the video, focusing on extracting key frames that best represent the overall narrative. The output should be a list of timestamps in seconds and their corresponding salient scores. \\ 
Output Format & The key timestamps are in the \colorbox{yyellow}{<timestamp\_1>}, \colorbox{yyellow}{<timestamp\_2>}, $\cdots$ seconds, Their saliency scores are  \colorbox{oorange}{<score\_1>}, \colorbox{oorange}{<score\_2>}, $\cdots$.  \\ 
Output Example & The key timestamps are in the 8.5, 10.0, 11.0, 12.0, 23.5, 44.5, 45.0 seconds. Their saliency scores are 1.8, 3.7, 4.5, 4.2, 2.1, 4.7, 4.2. \\ \midrule
\rowcolor{dt!50}
\multicolumn{2}{c}{Video Highlight Detection} \\
Instruction Example &  Watch the provided video and mark out the scenes that stand out based on the description: \colorbox{ppurple}{<query\_placeholder>}. Document the timestamps of these highlights and evaluate their saliency scores. \\ 
Output Format & There are \colorbox{ggrey}{<highlight\_moments\_number>} highlight moments in the \colorbox{yyellow}{<timestamp\_1>}, \colorbox{yyellow}{<timestamp\_2>}, $\cdots$ seconds, Their saliency scores are  \colorbox{oorange}{<score\_1>}, \colorbox{oorange}{<score\_2>}, $\cdots$.  \\ 
Output Example & There are 16 highlight moments in the 44.0, 46.0, 48.0, 50.0, 52.0, 54.0, 56.0, 58.0, 60.0, 62.0, 64.0, 66.0, 68.0, 70.0, 72.0, 74.0 second. Their saliency scores are 2.7, 4.0, 3.7, 3.3, 2.7, 3.0, 3.0, 3.0, 3.0, 3.0, 3.0, 3.0, 2.7, 3.0, 3.0, 3.0. \\ \midrule
\rowcolor{dt!50}
\multicolumn{2}{c}{Transcribed Speech Generation} \\
Instruction Example   &  Watch the video, transcribe the speech, and indicate when each segment starts and ends. \\
Output Format & Transcribed speech: \colorbox{yyellow}{<timestamp\_start>} - \colorbox{yyellow}{<timestamp\_end>} seconds, \colorbox{ggreen}{<transcribed\_speech>}. $\cdots$  \\ 
Output Example & Transcribed speech: 4.0 - 9.3 seconds, Dolby as well as we had over 7.7 million minutes viewed. This week we visit restaurant. 9.3 - 15.4 seconds, August by Chef John Besh in New Orleans 2015. Restaurant August is currently regarded as New. $\cdots$ \\
\bottomrule
\end{tabular}
\end{adjustbox}

\caption{Instruction template examples and formatted output answer for each task.}
\label{table:dataformat}
\end{table*}

\section{Instructions for Each Task}
\label{sec:ins4task}
The quality and diversity of instructions are essential in the construction process. We manually write well-designed instructions for each task as a good starting. Then we utilize GPT-4~\citep{OpenAI2023GPT4TR} to extend more diverse and flexible expressions based on the manual initialization. Eventually, we manually select and refine the LLM-generated instructions to obtain the final version. Inspired by the observation in $\text{M}^3\text{IT}$~\cite{Li2023M3ITAL} that using around five instructions per task is sufficient, we generate six high-quality instructions for each task. 
Tab.~\ref{table:dataformat} shows instruction template examples and formatted output answers for each task.

\section{Contribution Analysis of Each Task to Model Performance}
\label{sec:contribution-analysis}
\begin{table*}[]
\centering
\begin{adjustbox}{max width=\linewidth}
\begin{tabular}{@{}llll|ll|ll@{}}
\toprule
\multirow{3}{*}{Tasks in TimeIT} & \multicolumn{3}{c|}{Dense Captioning} & \multicolumn{2}{c|}{Highlight Detection} & \multicolumn{2}{c}{Temporal Grounding} \\
       & \multicolumn{3}{c|}{YouCook2}                 & \multicolumn{2}{c|}{QVHighlights}               & \multicolumn{2}{c}{Charades-STA}              \\ \cmidrule(l){2-8} 
      & {\small SODA\_c}   & {\small CIDEr}   & {\small F1}          & mAP                  & HIT@1        & 
                 R@1{\tiny (IoU=0.5)}          
                 & 
                 R@1{\tiny (IoU=0.7)} 
                 \\ \midrule
DVC+TVG & \colorbox{wwhite}{0.6} & \colorbox{wwhite}{1.9} & \colorbox{wwhite}{~~5.9} & \colorbox{wwhite}{11.2} & \colorbox{wwhite}{15.5} & \colorbox{wwhite}{\textbf{34.9}} & \colorbox{wwhite}{13.6} \\
DVC+TVG+\textbf{SLC} & \colorbox{ggreen}{1.1} & \colorbox{ggreen}{3.2} & \colorbox{ggreen}{12.1} & \colorbox{ggreen}{11.8} & \colorbox{ggreen}{16.5} & \colorbox{rred}{32.7} & \colorbox{ggreen}{13.9} \\
DVC+TVG+SLC+\textbf{VS} & \colorbox{wwhite}{1.1} & \colorbox{rred}{3.0} & \colorbox{ggreen}{12.2} & \colorbox{ggreen}{13.0} & \colorbox{ggreen}{19.0} & \colorbox{ggreen}{33.2} & \colorbox{ggreen}{\textbf{14.3}} \\
DVC+TVG+SLC+VS+\textbf{TSG} & \colorbox{ggreen}{\textbf{1.2}} & \colorbox{ggreen}{\textbf{3.4}} & \colorbox{ggreen}{\textbf{12.6}} & \colorbox{ggreen}{\textbf{14.5}} & \colorbox{ggreen}{\textbf{23.9}} & \colorbox{rred}{32.2} & \colorbox{rred}{13.4} \\
\bottomrule
\end{tabular}
\end{adjustbox}
\caption{Contributions (\colorbox{ggreen}{positive}/\colorbox{rred}{negative}) of tasks in TimeIT to model performance. The tasks include Dense Video Captioning (DVC), Temporal Video Grounding (TVG), Step Localization and Captioning (SLC), Video Summarization (VS), and Transcribed Speech Generation (TSG).}
\label{table:timeit-data-contribution}
\end{table*}
We examine the impact of individual tasks within the TimeIT dataset on model performance. 
Initially, we construct the TimeIT dataset with only DVC and TVG tasks, then gradually integrating additional tasks such as SLC, VS, and TSG, to assess their influence on model performance. 
As shown in Tab.~\ref{table:timeit-data-contribution}, introducing similar tasks (e.g., SLC to DVC) yields a positive impact (e.g., increasing F1 score on YouCook2 from 5.9 to 12.1). Overall, all 6 tasks are beneficial.

\section{Hyper-parameters for Instruction Tuning}
\label{sec:hyper-parameters}
\begin{table}[t!]
\centering
\begin{adjustbox}{max width=\linewidth}

\begin{tabular}{lc}
\toprule
Hyper-parameter & Value \\ \midrule
Patch size & 14 $\times$ 14 \\
Frame resolution & 224 $\times$ 224 \\ \midrule
Fine-tuning epochs & 3 \\
Batch size & 32 \\
Learning rate & 3e-5 \\
Warm-up learning rate & 1e-6 \\
Weight decay & 0.05 \\ 
AdamW $\beta$ & (0.9, 0.999) \\ \midrule 
Window size $L_W$ & 32 \\
Stride $S$ & 32 \\
Number of video tokens per window $N_V$ & 32 \\
Number of input frames $T$ & 96 \\
Max text length & 2048 \\ \midrule
Number of layers in video Q-Former & 2 \\
Number of layers in image Q-Former & 12 \\
Hidden size of image/video Q-Former ($D_Q$) & 768 \\
Hidden size of LLaMA-2 ($D_{LLM}$) & 4096 \\
\bottomrule
\end{tabular}
\end{adjustbox}
\caption{Hyper-parameters for instruction tuning.}
\label{table:hyper-parameters}
\end{table}

\begin{table*}[]
\centering
\begin{adjustbox}{max width=\linewidth}

\begin{tabular}{@{}cccc|ccc@{}}
\toprule
window size & stride & window overlap & \#video tokens & SODA\_c   & CIDEr  & F1 Score \\ \midrule
32 & 32 & \xmark & 96 & 2.9 & 9.6 & 19.0 \\ \midrule
32 & 16 & \cmark & 192 & 2.9 & 10.0 & 19.6 \\
16 & 16 & \xmark & 192 & \textbf{3.2} & \textbf{11.7} & \textbf{19.8} \\ \midrule
16 & 8 & \cmark & 384 & 3.1 & 10.8 & 19.5 \\
8 & 8 & \xmark & 384 & 3.1 & 11.2 & 19.7 \\
\bottomrule
\end{tabular}


\end{adjustbox}
\caption{Sliding window hyper-parameters sweep on YouCook2.}
\label{table:window-hyper-ablation}
\end{table*}

Tab.~\ref{table:hyper-parameters} lists hyper-parameters for instruction tuning. 
We also conduct an ablation on sliding window hyper-parameters. The results are on Tab.~\ref{table:window-hyper-ablation}. We adopt a window size$=$stride$=$32 for efficiency (higher compression rate~\citep{Ren2023TESTA} and fewer video tokens). However, a more thorough search may improve performance (window size$=$stride$=$16). Besides, non-overlapping windows outperform overlapping ones.

\begin{figure*}[t]
\centering
\includegraphics[width=\textwidth]{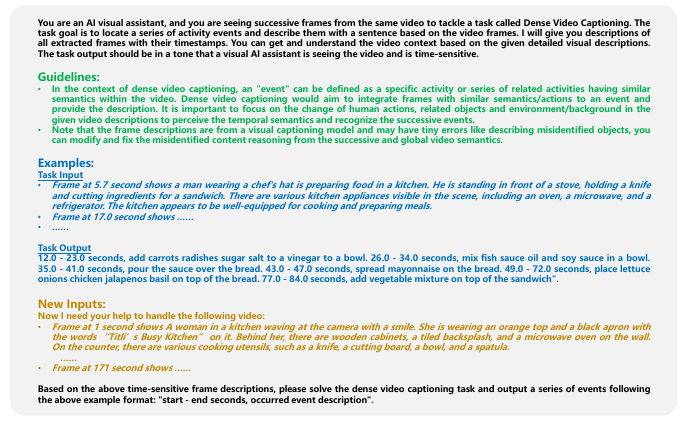}
\caption{ Examples of designed prompts for the InstructBLIP+ChatGPT pipeline. The input prompts encompass \textbf{(1)} the task definition, \textbf{(2)} specific guidelines, \textbf{(3)} an in-context example, and \textbf{(4)} the new instance input. The video information includes detailed frame descriptions from InstrucBLIP with the corresponding timestamp. We uniformly sample 50 frames for each video to obtain visual semantics.
}
\label{fig:chatgpt-prompt}
\end{figure*}

\begin{figure}[t]
\centering
\includegraphics[width=\linewidth]{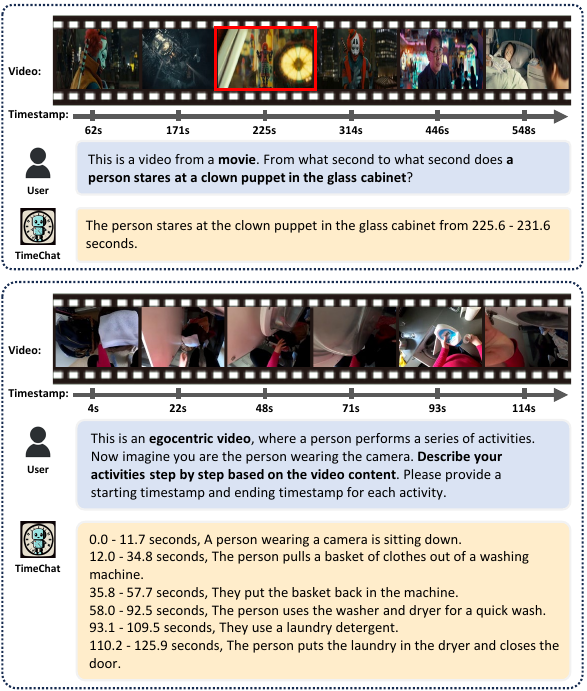}
\caption{Zero-shot transfer to new domains such as movie (upper) and egocentric videos (bottom).}
\label{fig:transfer-case}
\end{figure}

\begin{figure*}[t]
\centering
\includegraphics[width=\textwidth]{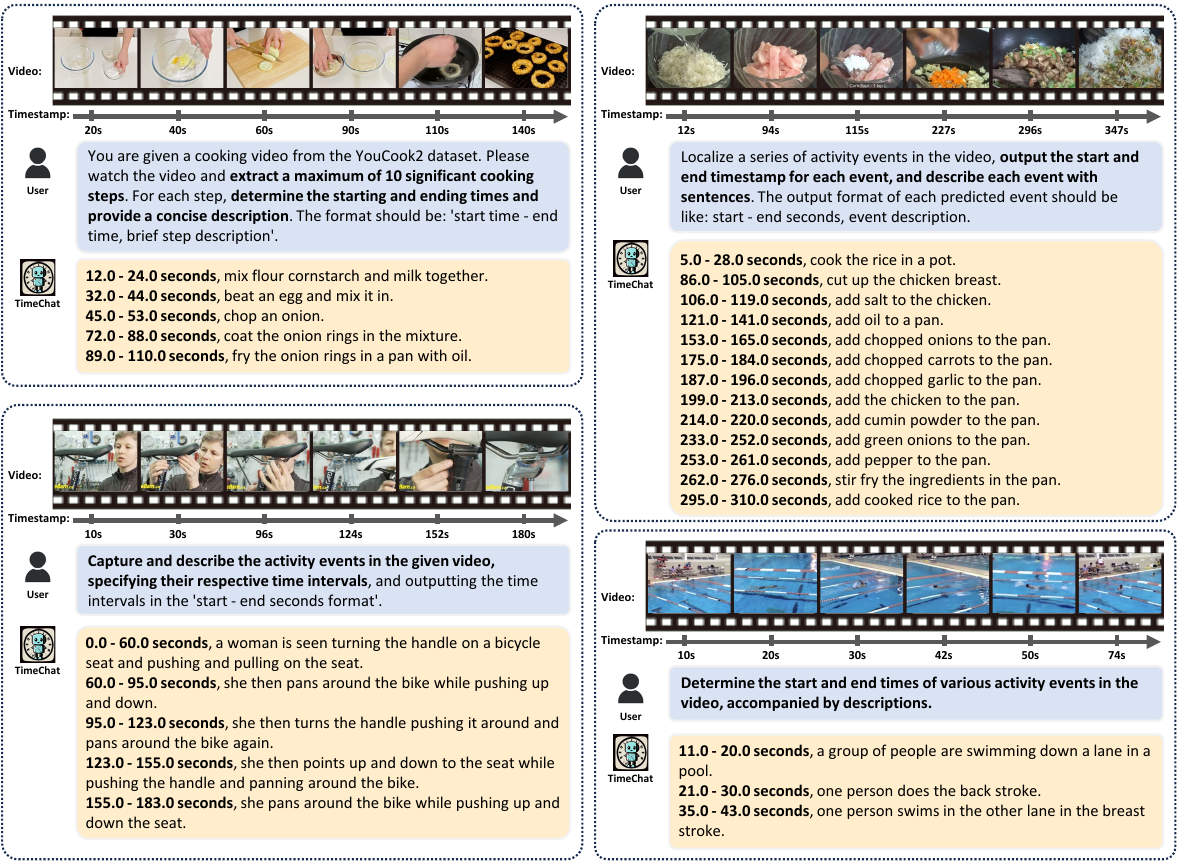}
\caption{Qualitative results on video dense captioning task. For each video, we ask \modelname to detect a series of events in the given video and output the corresponding timestamps and descriptions.
}
\label{fig:case_dvc}
\end{figure*}

\begin{figure*}[t]
\centering
\includegraphics[width=\textwidth]{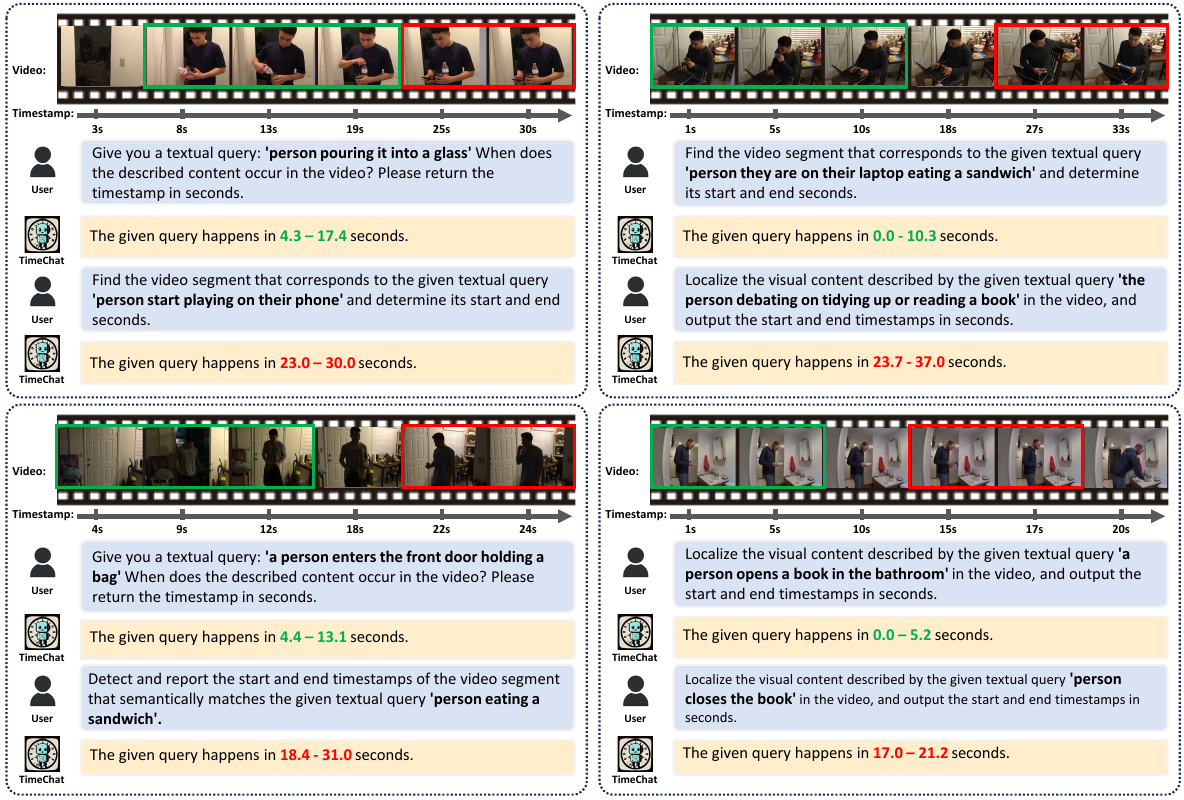}
\caption{Qualitative results for temporal video grounding task. For each video, we prompt our model to estimate the starting and ending timestamps for two specified queries (highlighted in bold). The predicted start and end timestamps and their corresponding segments are displayed in \textbf{\textcolor{green}{green}} and \textbf{\textcolor{red}{red}}, respectively.
}
\label{fig:case_tvg}
\end{figure*}

\begin{figure*}[t]
\centering
\includegraphics[width=\textwidth]{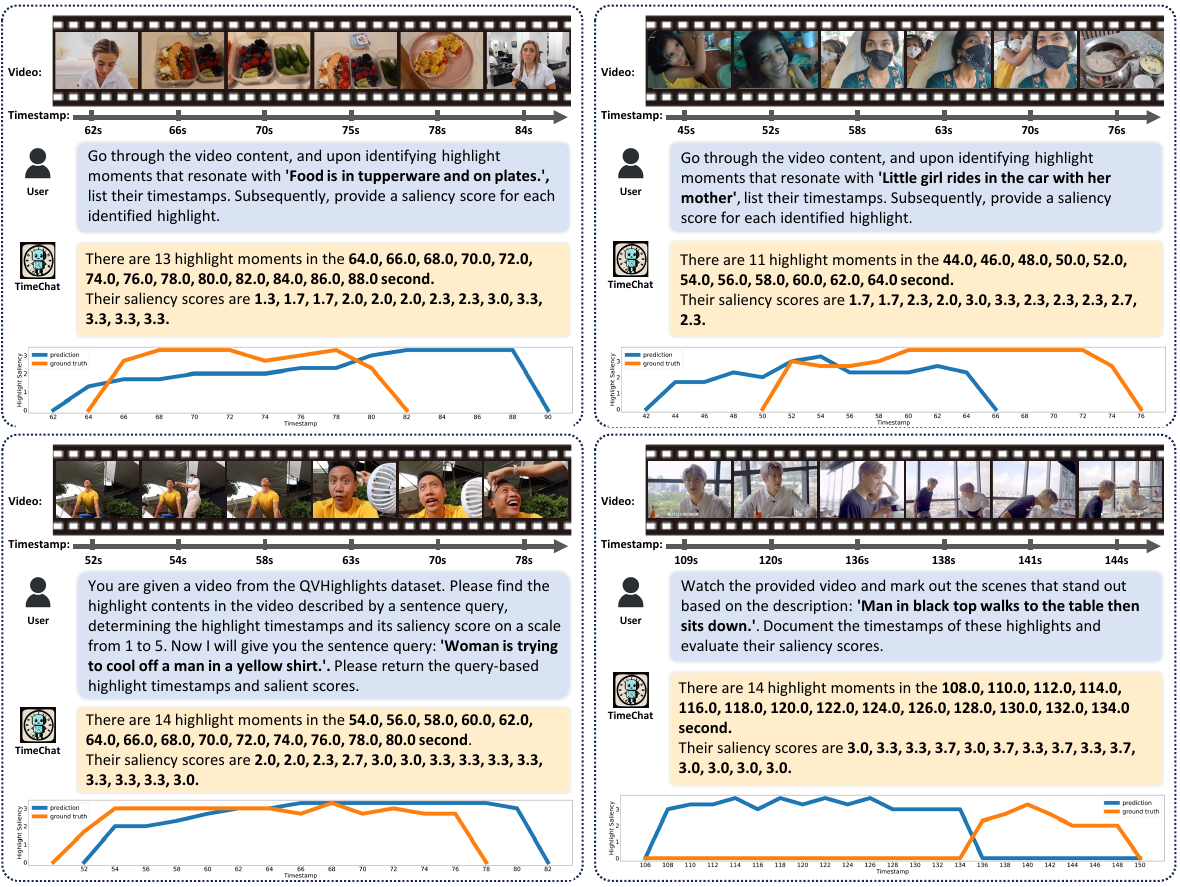}
\caption{Qualitative results for video highlight detection task. In each video, we instruct \modelname to pinpoint the most thrilling, remarkable, or emotive moment based on a specified query. The model is also required to assess the saliency score for each identified moment. We graph the saliency scores in relation to the moment's timestamp. The \textbf{\textcolor{orange}{orange}} curve denotes the ground truth, while the \textbf{\textcolor{blue}{blue}} curve signifies the predictions made by our model.
}
\label{fig:case_vhd}
\end{figure*}

\section{Details of Evaluation Datasets and Metrics}
\label{sec:evaluation}
TimeIT's 6 tasks can be grouped based on format similarity: \textbf{(1)} dense-captioning-formatted: DVC, SLC, and TSG; \textbf{(2)} highlight-detection-formatted: HD and VS; \textbf{(3)} temporal-grounding-formatted: TVG. For practicality and representation, we select the most relevant tasks from each group—DVC, HD, and TVG—for evaluation.

\textbf{(1) For dense video captioning}, we use the YouCook2 dataset~\citep{Zhou2017TowardsAL}, which has 1,790 untrimmed videos of cooking procedures. On average, each video lasts 320s and is annotated with 7.7 temporally-localized imperative sentences. The dataset is split into 1,333 videos for training and 457 videos for validation. We evaluate caption quality using CIDEr~\citep{Vedantam2014CIDErCI}. For an overall evaluation at the story level, we use the SODA\_c metric~\citep{Fujita2020SODASO}. We also report the F1 score, which is the harmonic mean of the average precision and recall across IoU thresholds of {0.3, 0.5, 0.7, 0.9}, to measure event localization performance.

\textbf{(2) For video highlight detection}, we use the QVHighlights dataset~\citep{Lei2021QVHighlightsDM}. It consists of over 10,000 videos annotated with human-written text queries. The evaluation metrics are mAP (mean average precision) with IoU thresholds of 0.5 and 0.75, and HIT@1 (the hit ratio of the highest-scored clip). 

\textbf{(3) For temporal video grounding}, we use the Charades-STA~\citep{Gao2017TALLTA} dataset. The dataset contains 6,670 videos and involves
16124 queries, where 12,404 pairs are used for training and 3,720 for testing. The average duration of the videos is 30.59 seconds and each video contains 2.41 annotated moments, and the moment has an average duration of 8.09 seconds. 
The evaluation metric is "R@1, IoU = $\mu$", which denotes the percentage of retrieved moments with an intersection over union (IoU) greater than $\mu$ compared to the ground truth, given language queries. 

\section{Details of Multi-model Pipelines}
\label{sec:pipeline}
We take VideoChat-Text~\citep{Li2023VideoChatCV} and InstructBLIP~\citep{Dai2023InstructBLIPTG}+ChatGPT~\citep{chatgpt} as the baselines of Multi-model Pipielines. These pipelines integrate specialized visual models with ChatGPT, which firstly convert video semantics into textual descriptions and then leverage ChatGPT to process all inputs to solve the target task.
\paragraph{VideoChat-Text} utilizes \texttt{ffmpeg} to extract key frames from the video at FPS=1. Then it leverages visual tools to obtain rich video information including action labels, frame summaries, video tags, comprehensive descriptions,
object positional coordinates, video narratives, timestamps, and segment-related details. The overall visual information will be processed by the ChatGPT to respond to user instructions. We design task-related prompts to endow VideoChat-Text with the capability to solve timestamp-sensitive tasks.
\paragraph{InstructBLIP+ChatGPT} endows a more powerful visual expert model, i.e. InstructBLIP, to describe each frame with exhaustive paragraphs containing detailed video semantics. We employ well-designed prompts (illustrated in \cref{fig:chatgpt-prompt}) for ChatGPT to solve each task. For video input, we uniformly sample 50 frames to obtain frame descriptions.

\section{Generalized to New Domains}
\label{sec:domain-generalize}
In Fig.~\ref{fig:transfer-case}, we show qualitative results in new domains such as movie~\cite{Yue2023Movie101AN} and egocentric videos~\cite{Grauman2021Ego4DAT}, demonstrating the generalization of \modelname to novel scenarios. This generalization is a key characteristic towards a practical video assistant and represents a fundamental difference between LLM-based \modelname and the current specialized models tailored for specific downstream datasets.

\section{More Qualitative Results}
\label{sec:more-res}
Within Figures~\ref{fig:case_dvc}-\ref{fig:case_vhd}, we present an extended range of qualitative results, encompassing dense video captioning, temporal video grounding, and video highlight detection tasks. Overall, our model demonstrates proficiency in executing a diverse array of intricate temporal localization tasks.

\end{document}